\documentclass[runningheads]{llncs}
\usepackage{graphicx}

\usepackage{tikz}
\usepackage{comment}
\usepackage{amsmath,amssymb} 
\usepackage{color}

\usepackage[accsupp]{axessibility}  

\usepackage[width=122mm,left=12mm,paperwidth=146mm,height=193mm,top=12mm,paperheight=217mm]{geometry}

\usepackage{xspace}
\usepackage[normalem]{ulem}

\newcommand{\shai}[2][]{%
    \ifthenelse{ \equal{#1}{} }
        {\textcolor{cyan}{(Shai) #2}}
        {\textcolor{cyan}{(Shai) \sout{#1\xspace}#2}}
}

\newcommand{\shir}[2][]{%
    \ifthenelse{ \equal{#1}{} }
        {\textcolor{blue}{(Shir) #2}}
        {\textcolor{blue}{(Shir) \sout{#1\xspace}#2}}
}

\newcommand{\yossi}[2][]{%
    \ifthenelse{ \equal{#1}{} }
        {\textcolor{orange}{(Yossi) #2}}
        {\textcolor{orange}{(Yossi) \sout{#1\xspace}#2}}
}

\newcommand{\tali}[2][]{%
    \ifthenelse{ \equal{#1}{} }
        {\textcolor{magenta}{(Tali) #2}}
        {\textcolor{magenta}{(Tali) \sout{#1\xspace}#2}}
}

\usepackage{dsfont}
\usepackage{xcolor}
\usepackage{tabularx}
\usepackage{sidecap}
\usepackage{arydshln} 
\usepackage{multirow}
\usepackage{rotating}
\usepackage{array}
\newcommand{\afterfigure}{\vspace*{-0em}}

\usepackage{pifont}
\definecolor{sup}{RGB}{154, 154, 184}  
\newcommand{\cls}{\texttt{[CLS]}}
\newcommand{\etal}{et al.}
\newcommand{\NN}{\mathsf{NN}}

\newcommand{\myparagraph}[1]{\paragraph{#1}}
\renewcommand{\afterfigure}{\vspace*{-0em}}

\begin{document}
\pagestyle{headings}
\mainmatter
\def\ECCVSubNumber{5}  

\title{Deep ViT Features as Dense Visual Descriptors}

\titlerunning{Deep ViT Features as Dense Visual Descriptors}
%
\author{Shir Amir\inst{1} \and Yossi Gandelsman\inst{2} \and Shai Bagon\inst{1} \and Tali Dekel\inst{1}}
\authorrunning{Amir et al.}
%
\institute{Dept. of Computer Science and Applied Math, The Weizmann Inst. of Science \and Berkeley Artificial Intelligence Research (BAIR)}
\maketitle
\begin{figure}[h]
\centering
\includegraphics[width=0.825\textwidth]{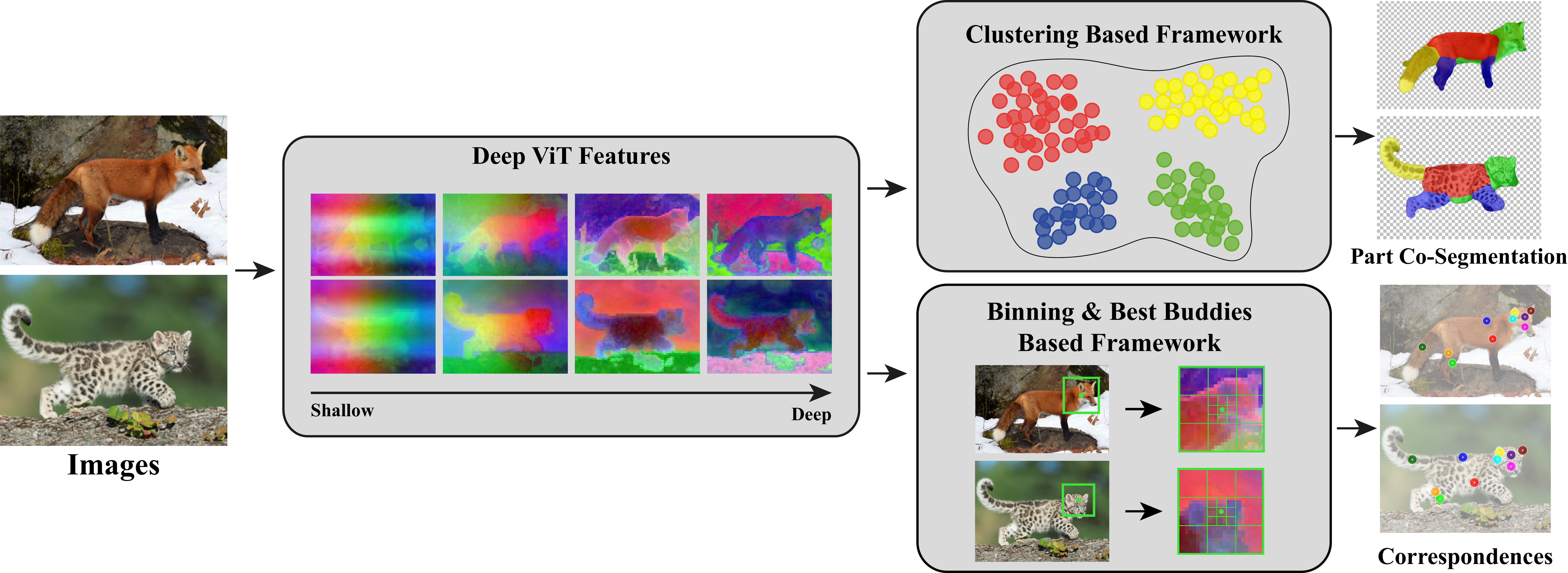}
\caption{Based on our new observations on deep ViT features, we devise \emph{lightweight zero-shot} methods to solve fundamental vision tasks (e.g. part co-segmentation and semantic correspondences). Our methods are applicable even in challenging settings where the images belong to different classes (e.g. fox and leopard).}
\label{fig:teaser}
\end{figure}

\begin{abstract}
We study the use of deep features extracted from a pre-trained Vision Transformer (ViT) as dense visual descriptors. We observe and empirically demonstrate that such features, when extracted from a self-supervised ViT model (DINO-ViT), exhibit several striking properties, including: (i) the features encode powerful, well-localized semantic information, at high spatial granularity, such as object \emph{parts}; (ii) the encoded semantic information is \emph{shared across related, yet different object categories}, and (iii) positional bias changes gradually \emph{throughout the layers}. These properties allow us to design simple methods for a variety of applications, including co-segmentation, part co-segmentation and semantic correspondences. To distill the power of ViT features from convoluted design choices, we restrict ourselves to \emph{lightweight zero-shot} methodologies (e.g., binning and clustering) applied directly to the features. Since our methods require no additional training nor data, they are readily applicable across a variety of domains. We show by extensive qualitative and quantitative evaluation that our simple methodologies achieve competitive results with recent state-of-the-art \emph{supervised} methods, and outperform previous unsupervised methods by a large margin. Code is available in \texttt{\url{dino-vit-features.github.io}}.

\end{abstract}

\section{Introduction}

``Deep Features" -- features extracted from the activations of layers in a pre-trained neural network -- have been extensively used as visual descriptors in a variety of visual tasks, yet have been mostly explored for CNN-based models. For example, deep features extracted from CNN models that were pre-trained for visual classification (e.g., VGG~\cite{vgg})  have been utilized  in numerous visual tasks including image generation and manipulation, correspondences, tracking and as a general perceptual quality measurement. 

Recently, Vision Transformers (ViT)~\cite{vit} have emerged as a powerful alternative architecture to CNNs. ViT-based models achieve impressive results in numerous visual tasks, while demonstrating better robustness to occlusions, adversarial attacks and texture bias compared to CNN-based models~\cite{naseer2021intriguing}. This raises the following questions: Do these properties reflect on the internal representations learned by ViTs? Should we consider deep ViT features as an alternative to deep CNN features? 
Aiming to answer these questions, we explore the use of deep ViT features as general dense visual descriptors: we empirically study their unique properties, and demonstrate their power through a number of real-world visual tasks. 

In particular, we focus on two pre-trained ViT models: a supervised ViT, trained for image classification~\cite{vit}, and a self-supervised ViT (DINO-ViT), trained using a self-distillation approach~\cite{dino}. In contrast to existing methods, which mostly focus on the features from the deepest layer~\cite{dino,simeoni2021localizing,wang2022tokencut}, we dive into the self-attention modules, and consider the various facets (tokens, queries, keys, values) \emph{across different layers}. We observe and empirically demonstrate that DINO-ViT features: (i)~encode powerful high-level information at high spatial resolution, i.e., capture semantic object \emph{parts},
(ii)~this encoded semantic information is \emph{shared across related, yet different object classes}, and (iii)~positional information gradually decreases \emph{throughout} layers, thus the intermediate layers encode position information as well as semantics. 
We demonstrate that these properties are not only due to the ViT architecture but also significantly influenced by the training supervision. 

Relying on these observations, we unlock the effectiveness of DINO-ViT features by considering their use in a number of fundamental vision tasks: co-segmentation, part co-segmentation, and semantic point correspondences. Moreover, equipped with our new observations, we tackle the task of part co-segmentation in a \emph{challenging unconstrained setting} where neither the number of input images, nor their domains are restricted. We further present how our part co-segmentation can be applied to videos.
To the best of our knowledge, we are the first to show results of part co-segmentation in such challenging cases (Fig.~\ref{fig:part_coseg_small_sets}).
We apply \emph{simple, zero-shot} methodologies to deep ViT features for all these tasks, which do not require further training. Deliberately avoiding large-scale learning-based models showcases the effectiveness of the learned DINO-ViT representations.   
We demonstrate that without bells and whistles, DINO-ViT features are already powerful enough to achieve competitive results compared to state-of-the-art models specifically designed and trained for each individual task. We thoroughly evaluate our performance qualitatively and quantitatively.

To conclude, our key contributions are: 
(i)~We uncover surprising \emph{localized semantic information}, far beyond saliency, readily available in ViT features.
(ii)~Our new observations give rise to \emph{lightweighted zero-shot} methodologies for tackling co- and part co-segmentation as well as semantic correspondences. 
(iii)~We are the first to show part co-segmentation in \emph{extreme settings}, showing how objects can be consistently segmented into parts across different categories, and across a variety of image domains, for some of which training data is scarce. 

%

\section{Related Work}\label{sec:related}

\myparagraph{CNN-based Deep Features.} 
Features of pre-trained CNNs are a cornerstone for various vision tasks
from object detection and segmentation \cite{RCNN,deeplab}, to image generation \cite{Shocher_2020_CVPR,Gatys_2016_CVPR}.
These representations were shown to align well with human perception \cite{Gatys_2016_CVPR,PerceptualLoss,LPIPS,mechrez2018contextual} and to encode a wide range of visual information - from low level features (e.g. edges and color) to high level semantic features (e.g. object parts)~\cite{olah2017feature,carter2019activation}. Nevertheless, they exhibit a strong bias towards texture~\cite{CNNtextureBias}, and lack positional information due to their shift equivariance~\cite{zhang2019shift_equ}. 
Moreover, their restricted receptive field~\cite{luo2016effective_rf} makes them capture mostly local information and ignore long-range dependencies~\cite{Wang_2018_CVPR}.
Here, we study the deep features of a less restrictive architecture - the Vision Transformer, as an alternative.

\myparagraph{Vision Transformer (ViT).}
 Vision Transformers~\cite{vit} have recently been used as powerful CNN alternatives. ViT-based models achieve impressive results in a variety of visual tasks \cite{vit,cat,detr}, while demonstrating better robustness to occlusions, adversarial attacks, and texture bias compared to CNN-based models~\cite{naseer2021intriguing}. 

In particular, Caron~\etal~\cite{dino} presented DINO-ViT -- a ViT model trained without labels, using a self-distillation approach. They observed that the attention heads of this model attend to salient foreground regions in an image. They further showed the effectiveness of DINO-ViT features for several tasks that benefit from this property, including image retrieval and object segmentation. 

Recent works follow this observation and utilize these features for object discovery~\cite{simeoni2021localizing,wang2022tokencut}, semantic segmentation~\cite{hamilton2022unsupervised} and category discovery~\cite{vaze2022generalized}.
All these works treat pre-trained DINO-ViT as a black-box, only considering features extracted from it's last layer, and their use as global or figure/ground-aware representations. 
In contrast, we examine the continuum of Deep ViT features \emph{across layers}, and dive into the different representations inside each layer (e.g. the keys, values, queries of the attention layers). We observe \emph{new} properties of these features besides being aware to foreground objects, and put these observations to use by solving fundamental vision tasks.


Concurrently, \cite{Raghu2021,cordonnier2019relationship,naseer2021intriguing} study theoretical aspects of the underlying machinery, aiming to analyze how ViTs process visual data compared to CNN models. Our work aims to bridge the gap between better understanding Deep ViT representations and their use in real-world vision tasks in a zero-shot manner.

\myparagraph{Co-segmentation.}
Co-segmentation aims to jointly segment objects common to all images in a given set.
Several unsupervised methods used hand-crafted descriptors \cite{cosegFaktor,Rubinstein13Unsupervised,regionMatchingCoSeg} for this task. Later, CNN-based methods applied supervised training \cite{deepObjectCoseg} or fine-tuning \cite{SSNM,cosegGroupwise,cycleSegNet} on \textit{intra-class} co-segmentation datasets. 
The supervised methods obtain superior performance, yet their notion of ``commonality" is restricted by their training data. Thus, they struggle generalizing to new \textit{inter-class} scenarios.
We, however, show a \emph{lightweight unsupervised} approach that is competitive to \emph{supervised} methods for intra-class co-segmentation \textit{and} outperforms them in the inter-class setting.

\myparagraph{Part Co-segmentation.} Given a set of images with similar objects, the task is to discover common object \emph{parts} among the images. Recent methods \cite{SCOPS,partcoseg21,choudhury21unsupervised} train a CNN encoder-decoder in a self-supervised manner to solve this task, while \cite{DFF} applies matrix factorization on pre-trained deep CNN features. 
In contrast, we utilize a pre-trained self-supervised ViT to solve this task, and achieve competitive performance to the methods above. Due to the zero-shot nature of our approach, we are able to apply part co-segmentation \emph{across classes}, and on domains that lack training supervision (see Fig.~\ref{fig:part_coseg_small_sets}). To the best of our knowledge, we are the first to address such challenging scenarios. 

\myparagraph{Semantic Correspondences.} 
Given a pair of images, the task is to find semantically corresponding points between them. Aberman \etal~\cite{aberman2018neural}~propose a sparse correspondence method for inter-class scenarios leveraging pre-trained CNN features.
Recent \emph{supervised} methods employ transformers for dense correspondence in images from the same scene \cite{sun2021loftr,COTR}. Cho \etal~\cite{cat} use transformers for semantic point correspondences by training directly on annotated point correspondences. We show that utilizing ViT features in a \emph{zero-shot} manner can be competitive to \emph{supervised} methods while being more robust to different pose and scale than previous unsupervised methods. 

\begin{figure}[t!]
  \centering
  \includegraphics[width=\textwidth]{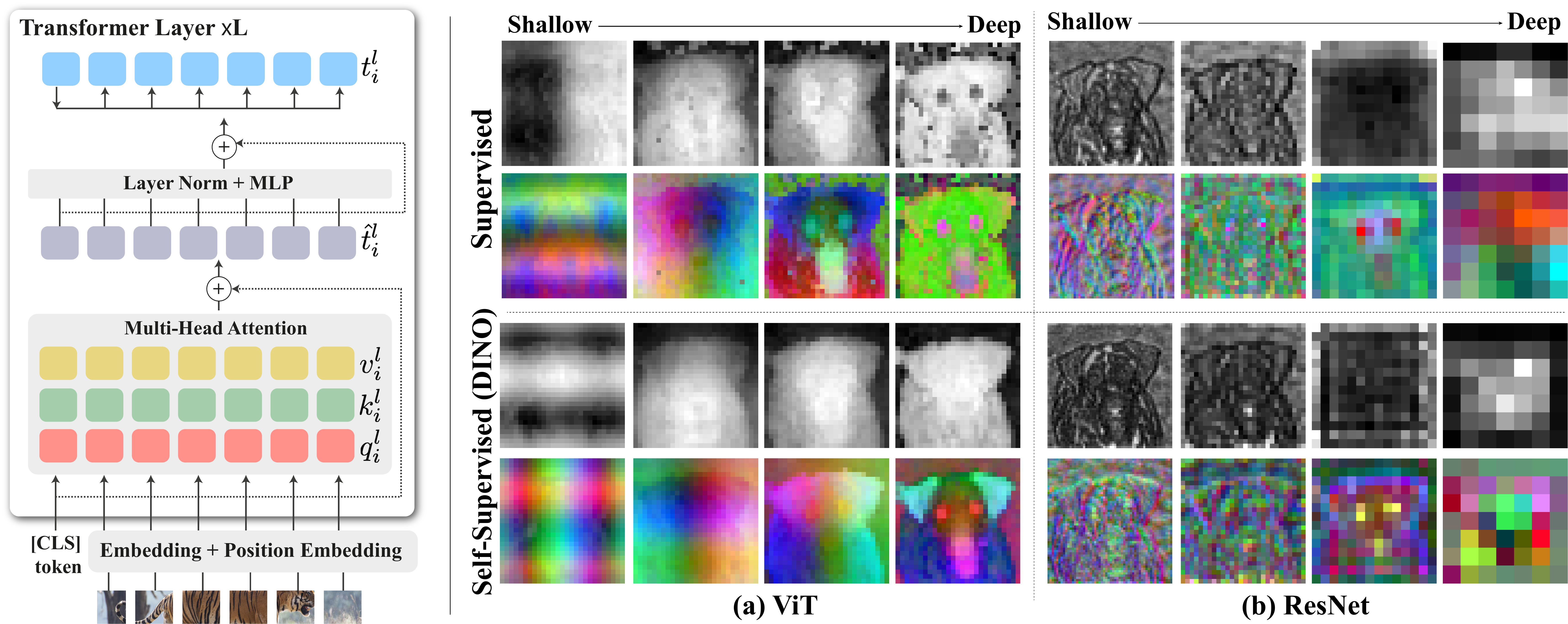} 
  \caption{\textit{ViT Architecture (Left).} An image is split into $n$ non-overlapping patches and gets a \cls token. These patches are embedded, added positional embeddings and passed through transformer layers. Each patch is directly associated with a set of features in each layer: a key, query, value and token; each can be used as patch descriptors. \textit{Deep features visualization via PCA (Right):} Applied on supervised and self-supervised (a) ViTs and (b) CNN-ResNet models. We fed 18 images from AFHQ~\cite{AFHQ} to each model, extract features from a given layer, and perform PCA on them. For each model, we visualize PCA components at each layer, for an example image (Dalmatian dog in Fig.~\ref{fig:co_parts_afhq} left): the first component is shown on the top, while second-to-fourth components are shown as RGB images below. ResNet PCA is upsampled for visualization purposes.
  }
  \label{fig:observations} \afterfigure
\end{figure}

\section{ViT Features as Local Patch Descriptors}\label{sec:vit}
\label{sec:vit}
We explore ViT features as \emph{local patch descriptors}.
In a ViT architecture, an image is split into $n$ non-overlapping patches $\left\{p_i\right\}_{i\in1..n}$ which are processed into \emph{spatial tokens} by linearly projecting each patch to a $d$-dimensional space, and adding learned positional embeddings. An additional \cls token is inserted to capture global image properties. The set of tokens are then passed through $L$ transformer encoder layers, each consists of normalization layers (LN), Multihead Self-Attention (MSA) modules, and MLP blocks (with skip connections):
{\small \begin{equation}
     \hat{T}^{l} = \mathsf{MSA}(\mathsf{LN}(T^{l-1}))  + T^{l-1},\;~~T^{l} = \mathsf{MLP}(\mathsf{LN}(\hat{T}^{l})) + \hat{T}^{l} 
\end{equation}}
where $T^{l}\!=\!\left[t_0^{l}, \dots, t_{n}^{l}\right]$ are the output tokens for layer $l$.

\noindent In each MSA block, tokens are linearly projected into queries, keys and values:

{\small \begin{equation}\label{eq:qkv}
    q^l_i = W_{q}^l\cdot t_i^{l-1},\;~~k^l_i = W_{k}^l\cdot t_i^{l-1},\;~~v^l_i = W_{v}^l\cdot t_i^{l-1}
\end{equation}}
\noindent which are then fused using multihead self-attention. Figure~\ref{fig:observations} Left illustrates this process, for full details see~\cite{vit}. Besides the initial image patches sampling, ViTs have no additional spatial sampling; hence, each image patch $p_i$ is \emph{directly} associated with a set of features: $\{q_i^l, k_i^l, v_i^l, t_i^l\}$, including its query, key, value and token, at each layer $l$, respectively. We next focus our analysis on using the \emph{keys} as `ViT features'. We justify this choice via ablation in Sections~\ref{sec:results_coseg} \& \ref{sec:results_correspondences}.

\subsection{Properties of ViT's Features}\label{sec:analysis}
 We focus on two pre-trained ViT models, both have the same architecture and training data, but differ in their training supervision: a \emph{supervised ViT}, trained for image classification using ImageNet labels~\cite{vit}, and a \emph{self-supervised ViT} (DINO-ViT), trained using a self-distillation approach~\cite{dino}. We next provide qualitative analysis of the internal representations learned by both models, and empirically originate their properties to the \emph{combination} of architecture  and  training supervision.
In Sec.~\ref{sec:results}, we show these properties enable several applications, through which we quantitatively validate our observations.

Figure~\ref{fig:observations} Right (a) shows a simple visualization of the learned representation by supervised ViT and DINO-ViT: for each model, we extract deep features (keys) from a set of layers, perform PCA, and visualize the resulting leading components. Figure~\ref{fig:observations} Right (b) shows the same visualization for two respective CNN-ResNet~\cite{Resnet} models trained using the same two supervisions as the ViT models: image classification, and DINO~\cite{dino}. This simple visualization shows fundamental differences between the internal representations of each model.

\begin{figure}[t!]
  \centering
  \includegraphics[width=\textwidth]{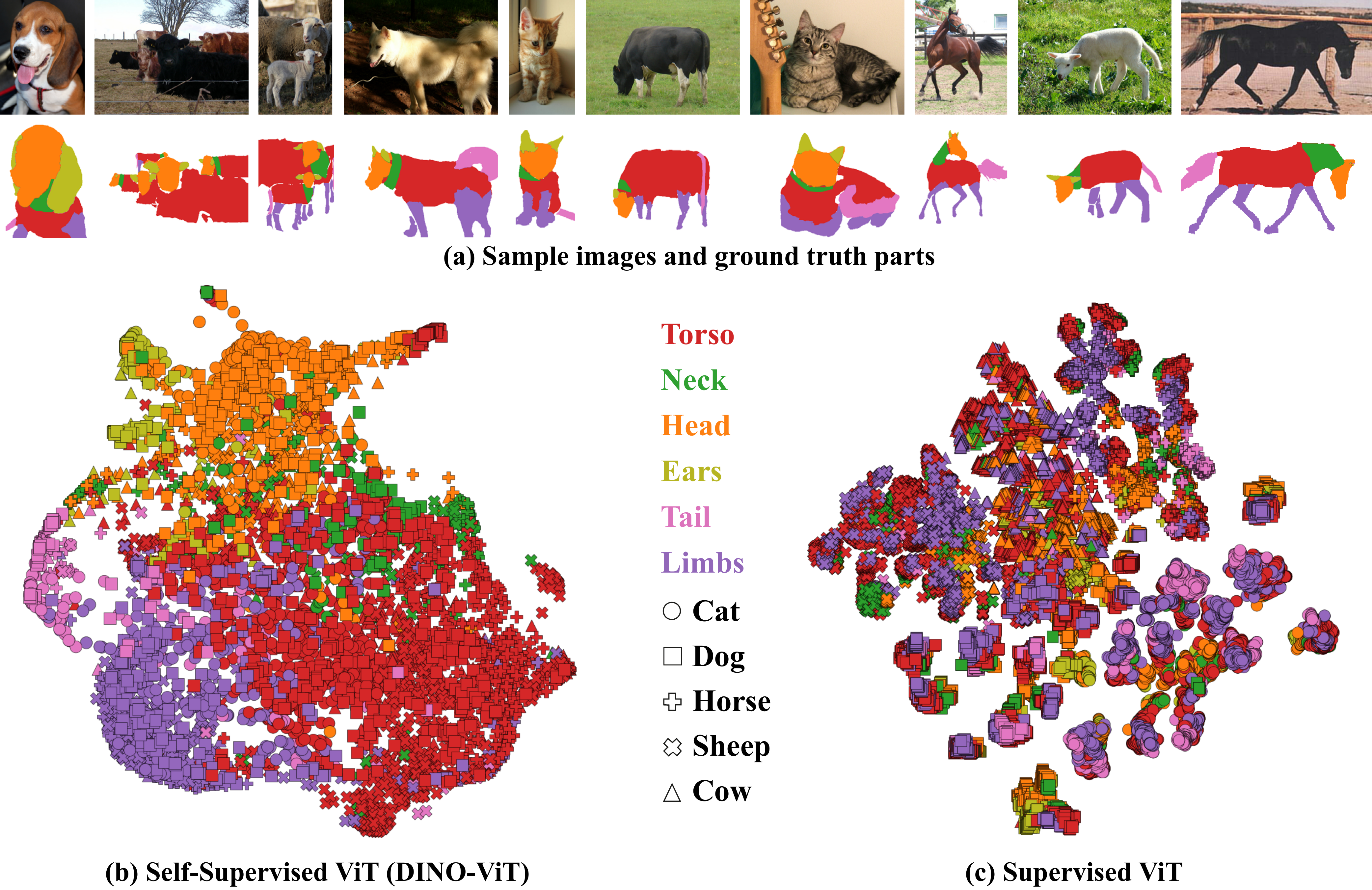} 
  \caption{\textit{t-SNE visualization.} We take 10 images from 5 animal categories from PASCAL-Parts~\cite{pascalparts}. (a) shows representative images and ground-truth part segments. 
  For each image we extract ViT features from DINO-ViT and a supervised ViT.
  For each model, all features are jointly projected to 2D using t-SNE~\cite{openTSNE}.
  Each 2D point is colored according to its ground-truth \emph{part}, while its shape represents the class.
  In (b) DINO-ViT features are organized mainly by parts, across different object categories, while in (c) supervised ViT features are grouped mostly by class, regardless of object parts.}
  \label{fig:t-sne} \afterfigure
\end{figure}

\myparagraph{Semantics vs. spatial granularity.} 
One noticeable difference between CNN-ResNet and ViT is that CNNs trade spatial resolution with semantic information in the deeper layers, as shown in Fig.~\ref{fig:observations} Right (b): the feature maps in the deepest layer have very low resolution ($\times{}32$ smaller than the input image), and thus provide poorly localized semantic information.
In contrast, ViT maintains the same spatial resolution through all layers. Also, the receptive field of ViT is the entire image in all layers~-- each token $t_i^l$ attends to all other tokens $t_j^l$. Thus, ViT features provide fine-grained semantic information \emph{and} higher spatial resolution.

\myparagraph{Representations across layers.} It is well known that the space of deep CNN-based features has a hierarchy of representation: early layers capture low-level elements such as edges or local textures (shallow layers in Fig.~\ref{fig:observations} Right (b)), while deeper layers gradually capture more high level concepts \cite{olah2017feature,carter2019activation,Shocher_2020_CVPR}. 
In contrast, we notice a different type of representation hierarchy in ViTs: \emph{Shallow features mostly contain positional information}, while in deeper layers, this is reduced in favor of more semantic features. For example, in Fig.~\ref{fig:observations} Right (a) the deep features distinguish dog features from background features, while the shallow features are gathered mostly based on their spatial location. Interestingly, intermediate ViT features contain both positional and semantic information.  

\myparagraph{Semantic information across super-classes.} Figure~\ref{fig:observations} Right (b) exhibits the supervised ViT model (top) produces ``noisier"  features compared to DINO-ViT (bottom). 
To further contrast the two ViTs, we employ t-SNE~\cite{openTSNE} to the keys of the last layer $\left[k_i^{11}\right]$, extracted from 50 animal images from PASCAL-Parts~\cite{pascalparts}.
Figure~\ref{fig:t-sne} presents the 2D-projected keys.
Intriguingly, the keys from a DINO-ViT show semantic similarity of body parts across different classes (grouped by \emph{color}), while the keys from a supervised ViT display similarity within each class regardless of body part (grouped by  \emph{shape}).
This demonstrates that while supervised ViT spatial features emphasize \emph{global} class information, DINO-ViT features have \emph{local} semantic information resembling semantic object \emph{parts}.

\begin{figure}[t!]
    \centering
    \includegraphics[width=0.8\linewidth]{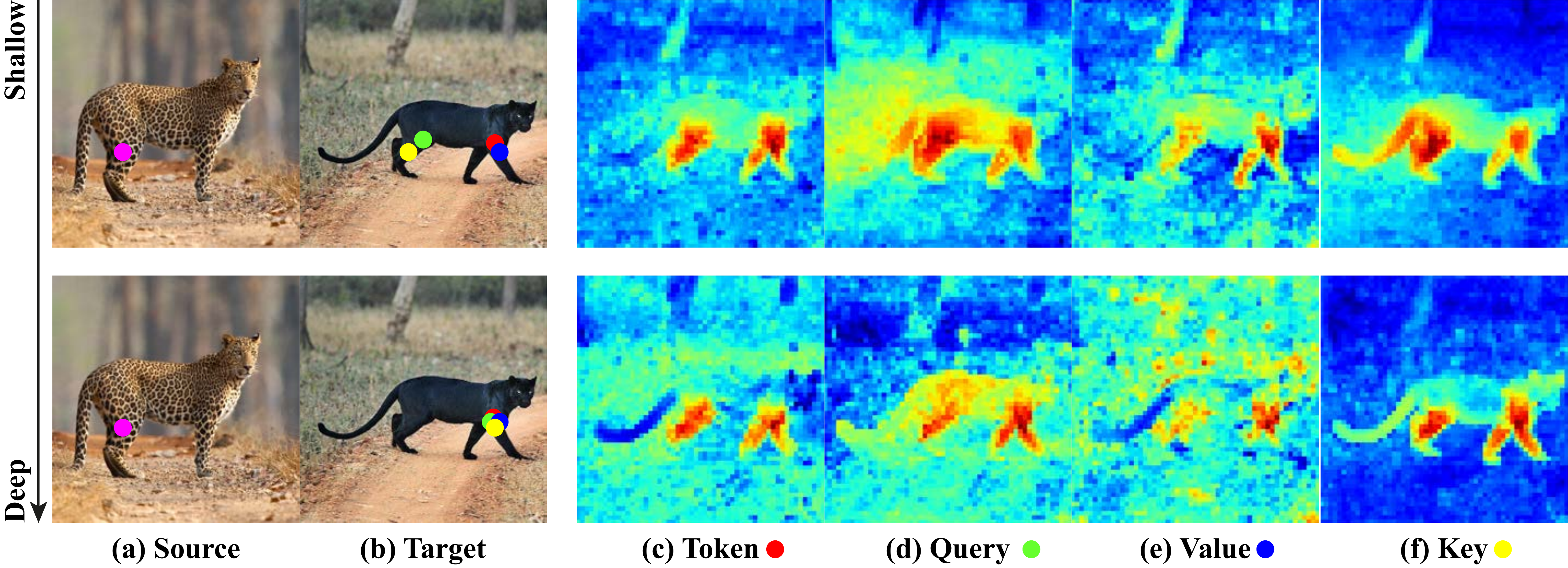}
    \caption{\textit{Facets of ViT:} We compute the similarity between a feature associated with the magenta point in the source image (a) to all features in the target image (b). We do this for intermediate features (top row) and features from the last layer (bottom row). (c-f) are the resulting similarity maps, when using different facets of DINO-ViT as features: tokens, queries, values and keys. Red indicates higher similarity. For each facet, the closest point in the target image is marked with a unique color, specified near the facet name. The keys (f) have cleaner similarity map compared to other facets. 
    }
    \label{fig:qkvt}\afterfigure
\end{figure}

\myparagraph{Different facets of ViT representation.}
So far, we focused on using keys as `ViT features'. However, ViT provides different facets that are also directly associated with each image patch (Fig.~\ref{fig:observations} Left). We empirically observe slight differences in the representations of ViT facets, as shown in Fig.~\ref{fig:qkvt}. In particular, we found the keys to provide a slightly better representation, e.g., they depict less sensitivity to background clutter than the other facets. In addition, both keys and queries posses more positional bias in intermediate layers than values and tokens. 

\begin{figure}[t!]
    \centering
    \includegraphics[width=\textwidth]{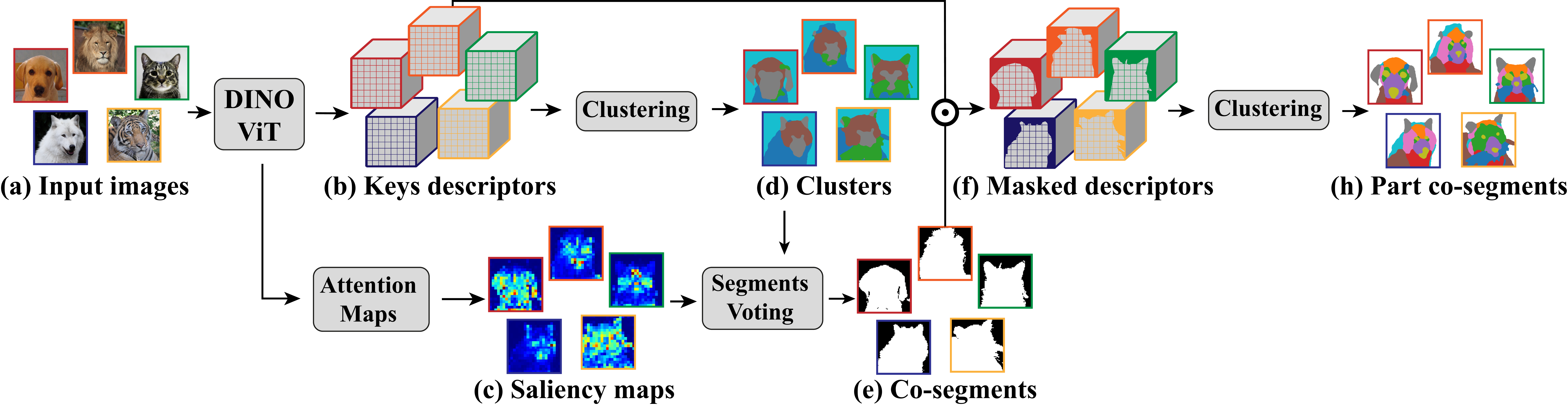}
    \caption{\textit{Co-segmentation \& part co-segmentation pipeline.} Input images (a) are fed separately to DINO-ViT to obtain (b) spatial dense descriptors and (c) saliency maps (from the ViT's self-attention maps). All the extracted descriptors are clustered together (d). Each cluster is assigned as foreground or background via a saliency maps based voting process. Foreground segments form the co-segmentation results (e). The process is repeated on foreground features (f) alone to yield the common parts (h).}
    \label{fig:coseg_pipeline} \afterfigure
\end{figure}

\section{Deep ViT Features Applied to Vision Tasks}\label{sec:applications}

We demonstrate the effectiveness of deep DINO-ViT features as local patch descriptors several visual tasks. We \emph{deliberately} apply only simple, lightweight methodologies on the extracted features, without any additional training nor fine-tuning, to showcase the effectiveness of DINO-ViT representations. For full implementation details, see supplementary material (SM).

\myparagraph{Co-segmentation.} Our co-segmentation approach, applied to a set of $N$ input images, comprises of two steps, followed by GrabCut~\cite{grabcut} to refine the binary co-segmentation masks, as  illustrated in Fig.~\ref{fig:coseg_pipeline}(a-e):

\begin{enumerate}
\setlength\itemsep{0mm}
    \item\emph{Clustering:} 
    We treat the set of extracted descriptors across all images and all spatial locations as a bag-of-descriptors, and cluster them using k-means. 
    At this stage, the descriptors are clustered into semantic common segments.
    As illustrated in Fig.~\ref{fig:observations} Right, the most prominent features' component distinguishes foreground and background, which ensures their separation. The result of this stage is $K$ clusters that induce segments in all images.
   
    \item\emph{Voting:} 
    We use a simple voting procedure to select clusters that are salient and common to most of the images. 
    Let $\mathsf{Attn}^\mathcal{I}_i$ be the mean \cls attention of selected heads in the last layer in image $\mathcal{I}$ of patch $i$. 
    Let $S^\mathcal{I}_k$ be the set of all  patches in image $\mathcal{I}$ belonging to cluster $k$.
    The saliency of segment $S^\mathcal{I}_k$ is:
    {\small \begin{equation}\label{eq:saliency}
        \mathsf{Sal}\left(S^\mathcal{I}_k\right) = \frac{1}{\left|S^\mathcal{I}_k\right|}\sum_{i\in S^\mathcal{I}_k} \mathsf{Attn}_i^\mathcal{I}
    \end{equation}}
    \noindent Each segment votes for the saliency of the cluster $k$:
    {\small \begin{equation}
    \mathsf{Votes}\left(k\right) = \mathds{1}_{\left[\sum_\mathcal{I}\texttt{Sal}\left(S^\mathcal{I}_k\right) \ge \tau \right]}
    \end{equation}}
For some threshold $\tau$. A cluster $k$ is considered ``foreground" iff its $\mathsf{Votes}\left(k\right)$ is above percentage $p$ of all the images.
\end{enumerate}

\myparagraph{Part Co-segmentation.} 
To further co-segment the foreground objects into common \emph{parts}, we  repeat the clustering step only on foreground descriptors, see Fig.~\ref{fig:coseg_pipeline}(f-h). By doing so, descriptors of common semantic parts across images are grouped together.
We further refine the part masks using multi-label CRF~\cite{krahenbuhl2011crf}. In practice, we found k-means to perform well, but other clustering methods (e.g. ~\cite{spectralclustering,spectraljitendra}) can be easily plugged in. 
For co-segmentation, the number of clusters is automatically set using the elbow method~\cite{ng2012elbow}, whereas for part co-segmentation, it is set to the desired number of object parts. Our method can be applied to a variety of object categories, and to arbitrary number of input images $N$, ranging from two to thousands of images. On small sets we apply random crop and flip augmentations for improved clustering stability (see SM for more details). 

\myparagraph{Point Correspondences.}
Semantic information is necessary yet insufficient for this task. For example, matching points on an animal's tail in Fig.~\ref{fig:teaser}, relying only on semantic information is ambiguous: all
points on the tail are equally similar. We reduce this ambiguity in two manners:
\begin{enumerate}
    \item \emph{Positional Bias:} We want the descriptors to be position-aware. 
    Features from earlier layers are more sensitive to their position in the image (see Sec.~\ref{sec:analysis});
    hence we use mid-layer features which provide a good trade-off between position and semantic information.
    \item \emph{Binning:} We incorporate context into each descriptor by integrating information from adjacent spatial features. This is done by applying log-binning to each spatial feature, as illustrated in Fig.~\ref{fig:teaser}. 
\end{enumerate}


\noindent{}To automatically detect reliable matches between images, we adopt the notion of ``Best Buddies Pairs`` (BBPs)~\cite{bestbuddies}, i.e., we only keep descriptor pairs which are mutual nearest neighbors.   
Formally, let $M=\{m_i\}$ and $Q=\{q_i\}$ be sets of binned descriptors from images $I_M$ and $I_Q$ respectively. 
The set of BBPs is thus:
{\small
\begin{equation}
    \mathsf{BB}(M, Q) = \left\{(m, q) ~|~ m\in M , ~q\in Q , \right.   \left.\NN(m,Q)=q\wedge\NN(q,M)=m\right\}
\end{equation}}
Where $\NN(m,Q)$ is the nearest neighbor of $m$ in $Q$ under cosine similarity.

\myparagraph{Resolution Increase.}\label{sec:resolution_trick}
The spatial resolution of ViT features is inversely proportional to size of the \emph{non-overlapping} patches, $p_i$. 
Our applications benefit from higher spatial feature resolution.
We thus modify ViT to extract, at test time, \emph{overlapping} patches, interpolating their positional encoding accordingly.
Consequently, we get, without any additional training, ViT features at finer spatial resolution.
Empirically, we found this method to work well in all our experiments. 

\section{Results}\label{sec:results}

\subsection{Part Co-segmentation}\label{sec:results_part_coseg}


\begin{figure}[!t]
    \centering
    \includegraphics[width=\linewidth]{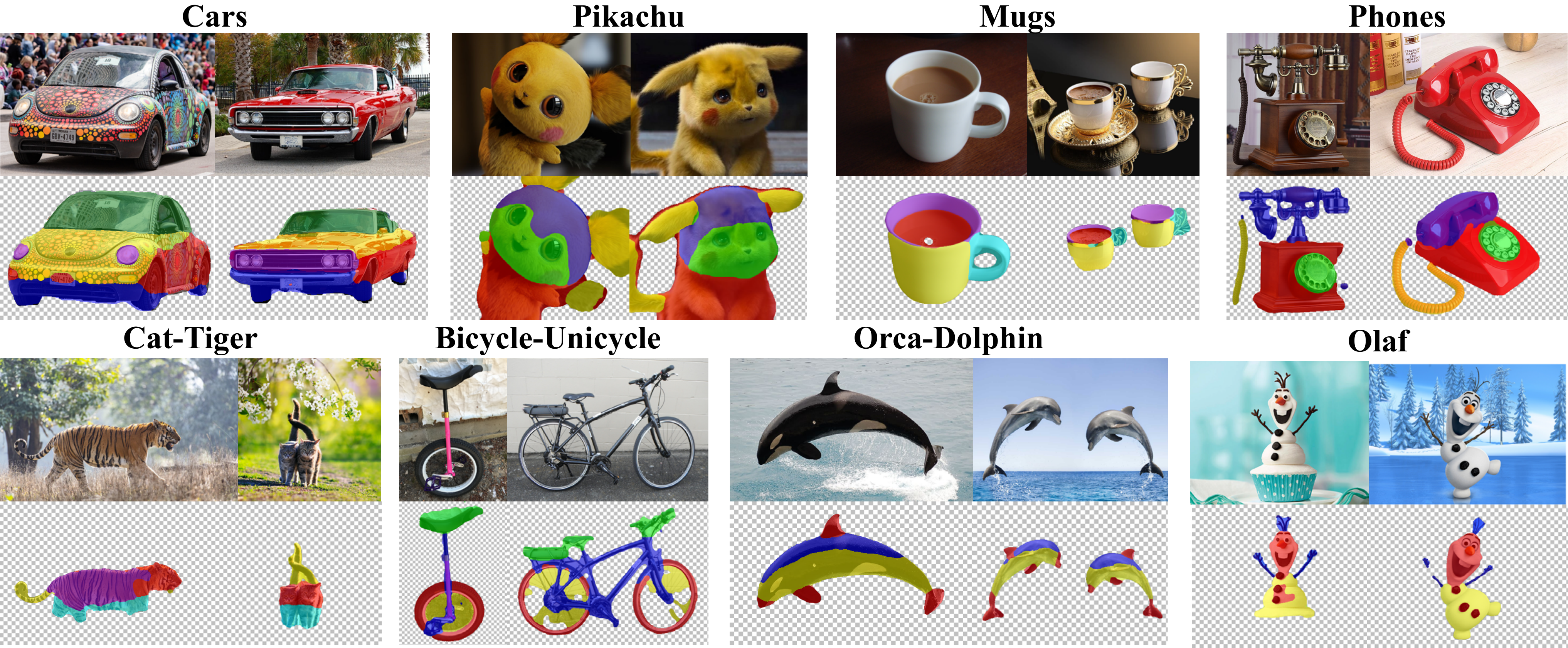}
    \caption{\textit{Part Co-segmentation of Image \emph{Pairs}:} Our method semantically co-segments common object parts given as little as two input images. See the SM for more examples.}
    \label{fig:part_coseg_small_sets}
\end{figure}

\begin{figure}[!t]
    \centering
    \includegraphics[width=\linewidth]{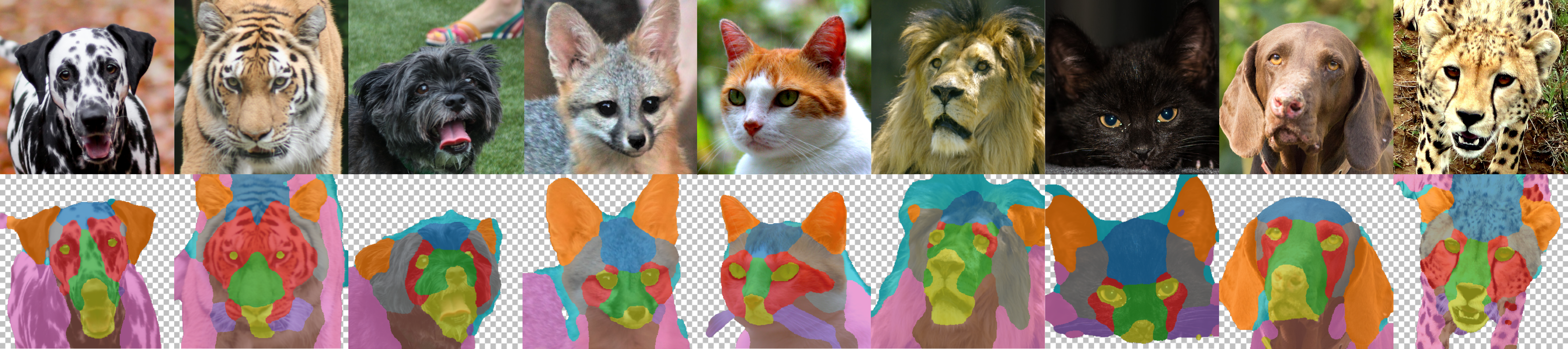}
    \caption{\textit{Part Co-segmentation on AFHQ:} We apply our method on the test set of AFHQ~\cite{AFHQ} containing 1.5K images of different animal faces. More results are in SM.}
    \label{fig:co_parts_afhq} \afterfigure
\end{figure}

\myparagraph{Challenging small sets.} In Fig.~\ref{fig:part_coseg_small_sets}, we present several image pairs collected from the web. These examples pose challenge due to different appearance (e.g. cars, phones), different classes (e.g. bicycle-unicycle, cat-tiger) and belonging to domains that are difficult to accommodate training sets for (e.g. pikachu, olaf). Our zero-shot method manages to provide semantically consistent part segments for each image pair. For example, in the bicycle-unicycle example the tires, spokes, chassis and saddle parts are consistently found. To the best of our knowledge, we are the first to handle such challenging cases.

\myparagraph{Video part Co-segmentation.} We extend our framework to work on videos by applying it to frames of a single video. Since DINO-ViT features are consistent across video frames, applying our observations to video co-segmentation yields temporally consistent parts. To the best of our knowledge, we are the first to apply part co-segmentation on videos. We include multiple examples in the SM. 

\myparagraph{Inter-class results.} In Fig.~\ref{fig:co_parts_afhq} we apply our part co-segmentation with $k=10$ parts on AFHQ~\cite{AFHQ} test set, containing 1.5K images of different animal faces. Our method provides consistent parts across \emph{different} animal classes, e.g. ears marked in orange, forehead marked in blue, whiskers marked in purple, etc. 

\begin{figure}[!t]

    \centering
    \includegraphics[width=\textwidth]{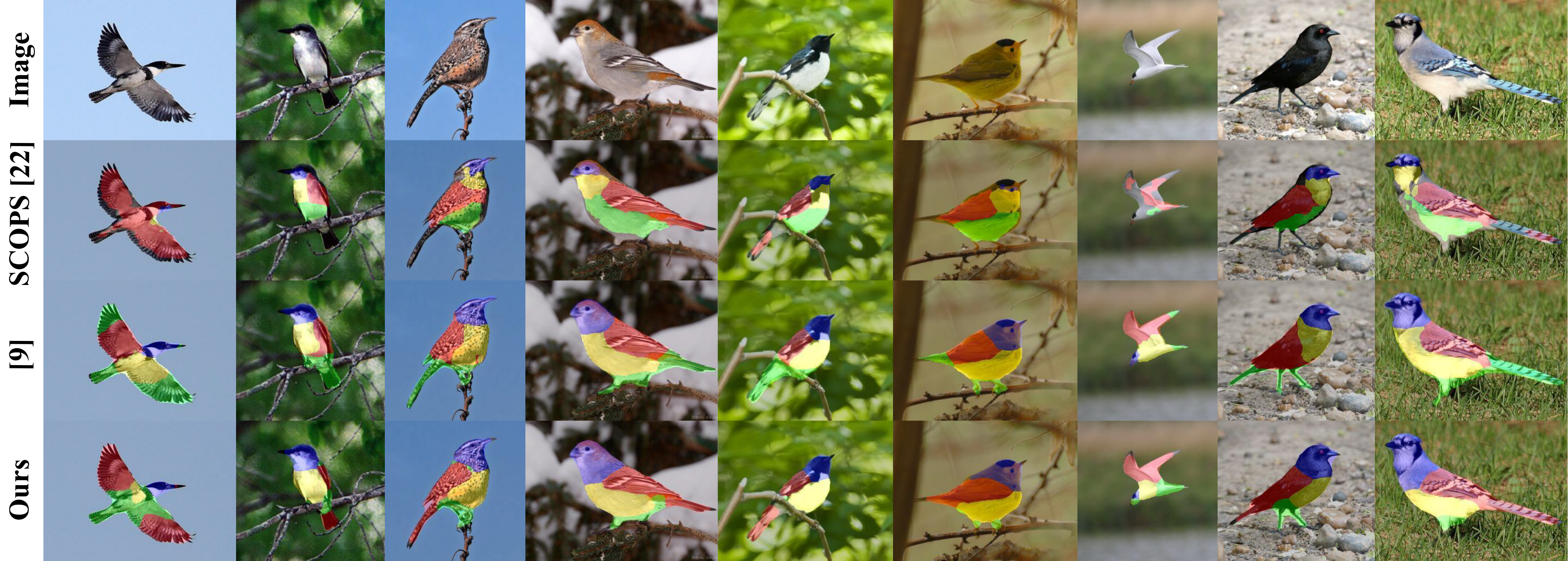}
    \caption{\textit{Part co-segmentation comparison on CUB:} We show results on randomly chosen images from CUB~\cite{CUB}. Our results are more semantically consistent across parts than the \emph{supervised} SCOPS~\cite{SCOPS} and are competitive to the \emph{supervised} Choudhury~\etal~\cite{choudhury21unsupervised}.}\label{fig:part_coseg_comp}
    \afterfigure
\end{figure}

\myparagraph{CUB~\cite{CUB} evaluation.}
Following~\cite{SCOPS,choudhury21unsupervised}, we evaluate performance on CUB~\cite{CUB} test set, which contains 5K images of different bird species. 
Following~\cite{SCOPS}, we measure the key-point regression error between the predicted and ground truth landmarks in Tab.~\ref{tab:part_coseg_lm} on three test sets from CUB~\cite{CUB}. In addition, we follow~\cite{choudhury21unsupervised} treating the part segments as clusters, and report NMI and ARI. FG-NMI and FG-ARI disregard the background part as a cluster.
Our method surpasses unsupervised methods \emph{by a large margin}, and is competitive to~\cite{choudhury21unsupervised} which is \emph{supervised} by foreground masks. Figure~\ref{fig:part_coseg_comp} shows our method produces more semantically coherent parts, with similar quality to~\cite{choudhury21unsupervised}. Further evaluation on the CelebA~\cite{CelebA} dataset is available in the SM.

\begin{table}[!t]
    \centering
    \begin{tabular}{l c c c c c c c}
    \hline
     \multirow{2}{*}{Method} & \multicolumn{3}{c}{\underline{key-point regression $\downarrow$}} & \multirow{2}{*}{FG-NMI $\uparrow$} & \multirow{2}{*}{FG-ARI $\uparrow$} & \multirow{2}{*}{NMI $\uparrow$} & \multirow{2}{*}{ARI $\uparrow$} \\
    
    & CUB-01 & CUB-02 & CUB-03 & & & & \\
    \hline
    \textit{supervised} & & & & & & & \\
    SCOPS \cite{SCOPS}$^\ddagger$ (model) & 18.3 & 17.7 & 17.0 & 39.1 & 17.9 & 24.4 & 7.1 \\
    Huang and Li \cite{Huang_2020_CVPR}$^\dagger$ & \underline{15.1} & 17.1 & \underline{15.7} & - & - & 26.1 & 13.2 \\
    Choudhury \etal \cite{choudhury21unsupervised}$^\ddagger$ & \textbf{11.3} & \underline{15.0} & \textbf{10.6} & \textbf{46.0} & \textbf{21.0} & \textbf{43.5} & \textbf{19.6} \\ 
    \hline
    \textit{unsupervised} & & & & & & & \\
    ULD \cite{ULD1,ULD2} & 30.1 & 29.4 & 28.2 & - & - & - & - \\
    DFF \cite{DFF} & 22.4 & 21.6 & 22.0 & 32.4 & 14.3 & 25.9 & 12.4 \\
    SCOPS \cite{SCOPS} (paper) & 18.5 & 18.8 & 21.1 & - & - & - & - \\
    \hline
    Ours & 17.1 & \textbf{14.7} & 19.6 & \underline{39.4} & \underline{19.2} & \underline{38.9} & \underline{16.1} \\

    \hline
    \end{tabular}
    \caption{\textit{Part Co-segmentation results:} We report mean error of landmark regression on three CUB~\cite{CUB} test sets, and NMI and ARI~\cite{choudhury21unsupervised} measures on the entire CUB test set. All methods predict $k=4$ parts. $\dagger$ method uses image-level supervision, $\ddagger$ methods use ground truth foreground masks as supervision.}
    \label{tab:part_coseg_lm}\afterfigure
\end{table}

\subsection{Co-segmentation}\label{sec:results_coseg}
\begin{figure}[!t]
    \centering
    \includegraphics[width=\textwidth]{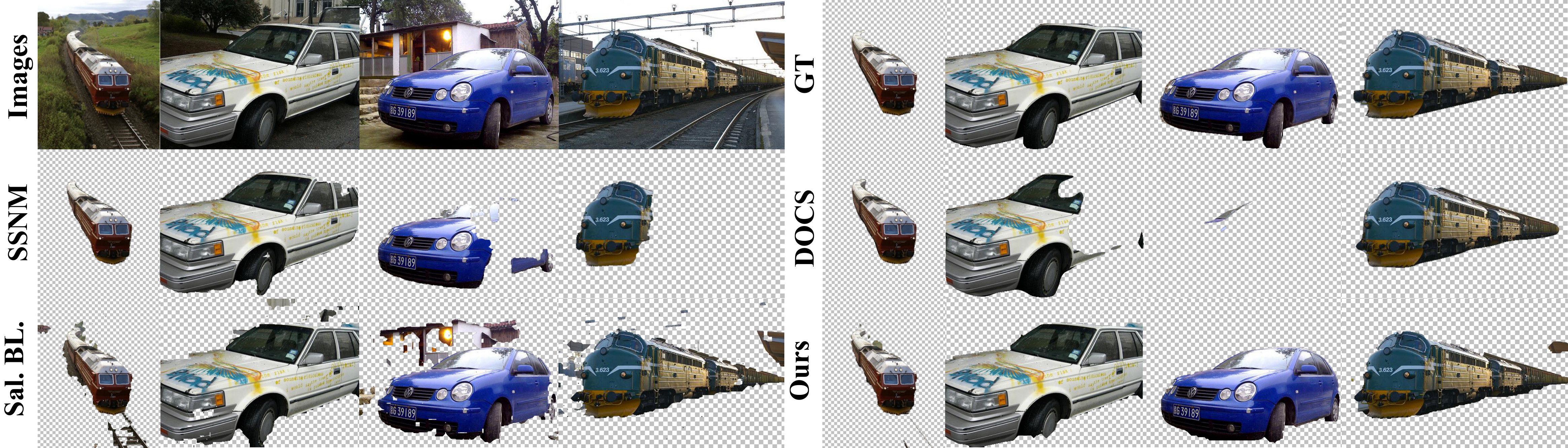}
    \caption{\textit{PASCAL-CO for inter-class co-segmentation:} Each set contains images from related classes. Our method captures regions of all common objects from different classes, contrary to supervised methods~\cite{deepObjectCoseg,SSNM}. Saliency Baseline~\cite{dino} results are noisy.}\label{fig:pascalco}
\end{figure}

We evaluate our performance on several \emph{intra-class} co-segmentation datasets of varying sizes - MSRC7~\cite{MSRC}, Internet300~\cite{Rubinstein13Unsupervised} and PASCAL-VOC~\cite{pascalvoc}.
Furthermore, to evaluate \emph{inter-class} co-segmentation, we compose a new dataset from PASCAL~\cite{pascalvoc} images, named ``PASCAL Co-segmentation'' (PASCAL-CO).
Our dataset has forty sets of six images, each from semantically related classes (e.g., car-bus-train, bird-plane). Fig.~\ref{fig:pascalco} shows a sample set, the rest is in the SM.

\begin{table}[!t]
    \centering
    \begin{tabular}{l| c| c| c| c| c| c| c| c| c}
    \hline
    Method & Training Set & \multicolumn{2}{c|}{\shortstack{ MSRC\\~\cite{MSRC}}} & \multicolumn{2}{c|}{\shortstack{Internet300\\~\cite{Rubinstein13Unsupervised}}} & \multicolumn{2}{c|}{\shortstack{PASCAL\\-VOC~\cite{pascalvoc}}} & \multicolumn{2}{c}{\shortstack{PASCAL\\-CO}}\\
    \cline{3-10}
    & & $\mathcal{J}_m$ & $\mathcal{P}_m$ & $\mathcal{J}_m$ & $\mathcal{P}_m$ & $\mathcal{J}_m$ & $\mathcal{P}_m$ & $\mathcal{J}_m$ & $\mathcal{P}_m$ \\
    \hline
    \textit{supervised} & & & & & & & & &\\
    SSNM \cite{SSNM} & COCO-SEG & 81.9 & 95.2 & 74.1 & 93.6 & 71.0 & \underline{94.9} & \underline{74.2} & \underline{94.5} \\
    DOCS \cite{deepObjectCoseg} & VOC2012 & 82.9 & 95.4 & 72.5 & 93.5 & \underline{65.0} & 94.2 & 34.9 & 53.7 \\
    CycleSegNet \cite{cycleSegNet} & VOC2012 & \textbf{87.2} & \textbf{97.9} & \underline{80.4} & - & \textbf{75.4} & \textbf{95.8} & - & - \\
    Li et al. \cite{cosegGroupwise} & COCO & - & - & \textbf{84.0} & \textbf{97.1} & 63.0 & 94.1 & - & - \\
    \hline
    \textit{unsupervised} & & & & & & & & &\\
    Hsu \etal \cite{hsu2018} & - & - & - & 69.8 & 92.3 & 60.0 & 91.0 & - & -  \\
    DeepCO3 \cite{hsu2019} & - & 54.7 & 87.2 & 53.4 & 88.0 & 46.3 & 88.5 & 37.3 & 74.1 \\
    TokenCut\cite{wang2022tokencut} & - & 81.2 & 94.9 & 65.2 & 91.3 & 57.8 & 90.6 & 75.8 & 93.0 \\
    Faktor \etal \cite{cosegFaktor} & - & 77.0 & 92.0 & - & - & 46.0 & 84.0 & 41.4 & 79.9 \\
    Rubinstein \etal \cite{Rubinstein13Unsupervised} & - & 74.0 & 92.2 & 57.3 & 85.4 & - & - & - & - \\
    \hline
    Ours & - & \underline{86.7} & \underline{96.5} & 79.5 & \underline{94.6} & 60.7 & 88.2 & \textbf{79.5} & \textbf{94.7} \\
    \hline
    \end{tabular}
    \caption{\textit{ Co-segmentation evaluation:} We report mean Jaccard index $\mathcal{J}_m$ and precision $\mathcal{P}_m$ over all sets in each dataset. We compare to unsupervised methods~\cite{cosegFaktor,Rubinstein13Unsupervised} and methods supervised with ground truth segmentation masks~\cite{SSNM,deepObjectCoseg,cycleSegNet,cosegGroupwise}.}\label{tab:coseg_table} \afterfigure
\end{table}

\begin{table}[!t]
    \centering
    \begin{tabular}{c| c c| c c| c| c| c| c}
    \hline
    & \multicolumn{2}{c|}{DINO Saliency Baselines} & \multicolumn{2}{c|}{Sup. Saliency Baselines} & \multicolumn{4}{c}{Ours}\\
    \hline 
    & \underline{ViT} & \underline{ResNet} & \underline{ViT} & \underline{ResNet} & Keys & Tokens & Queries & Values\\
    $\mathcal{J}_m$ & 75.0 & 37.7 & 39.9 & 40.0 & \textbf{79.5} & 69.2 & 72.7 & 49.2\\
    $\mathcal{P}_m$ & 93.1 & 78.1 & 69.7 & 78.9 & \textbf{94.7} & 90.68 & 91.7 & 83.3\\
    
    \hline
    \end{tabular}
    \caption{\textit{ Co-segmentation ablation:} 
    on PASCAL-Co for saliency baselines and our method using different ViT facets. Our method surpasses all baselines, and our choice of keys yields better performance than default chosen DINO-ViT tokens.}
    \label{tab:coseg_table_ablation} 
\end{table}


\myparagraph{Quantitative evaluation.} We compare our \emph{unsupervised} approach to state-of-the-art \emph{supervised} methods, trained on large datasets with ground truth segmentation masks, \cite{SSNM,deepObjectCoseg,cosegGroupwise,cycleSegNet}; and \emph{unsupervised} methods, \cite{cosegFaktor,Rubinstein13Unsupervised,wang2022tokencut,hsu2018,hsu2019}. We report Jaccard Index ($\mathcal{J}_m$), which reflects both precision (covering the foreground) and accuracy (no foreground ``leakage"), and mean precision ($\mathcal{P}_m$). The results appear in Table~\ref{tab:coseg_table}. Our method surpasses the unsupervised methods \emph{by a large margin}, and is competitive to the \emph{supervised} methods. In the \emph{inter-class} scenario (PASCAL-CO), our method surpasses all other methods.

\myparagraph{Ablation.} We conduct an ablation study to validate our observations in~\ref{sec:analysis}. As mentioned in Sec.~\ref{sec:related}, Caron \etal~\cite{dino} observed DINO-ViT attention heads attend to salient regions in the image, and threshold them to perform object segmentation. We name their method ``DINO-ViT Saliency Baseline'' as mentioned in Table~\ref{tab:coseg_table_ablation}. We apply the same baseline with attention heads from a supervised ViT (Sup. ViT Saliency Baseline).
To compare ViT with CNN representations, we also apply a similar method thresholding ResNet features (DINO / Sup. ResNet Saliency Baseline), implementation details are in the SM. We also ablate our method with different facets. The supervised ViT performs poorest, while both ResNet baselines perform similarly. The DINO-ViT baseline exceeds them and is closer to our performance. The remaining performance gap between our method and the DINO-ViT baseline can be attributed to one bias in the DINO-ViT baseline - it captures foreground salient objects regardless of their commonality to the other objects in the images. For example, the house behind the blue car in Fig.~\ref{fig:pascalco} is captured by DINO-ViT Saliency Baseline but is not captured by our method. This corroborates our observation that the properties of DINO-ViT stem from both architecture and training method. The facet ablation demonstrates our observation that keys are superior than other facets. 

\subsection{Point Correspondences}\label{sec:results_correspondences}

\begin{figure}[!t]
    \centering
    \includegraphics[width=1\textwidth]{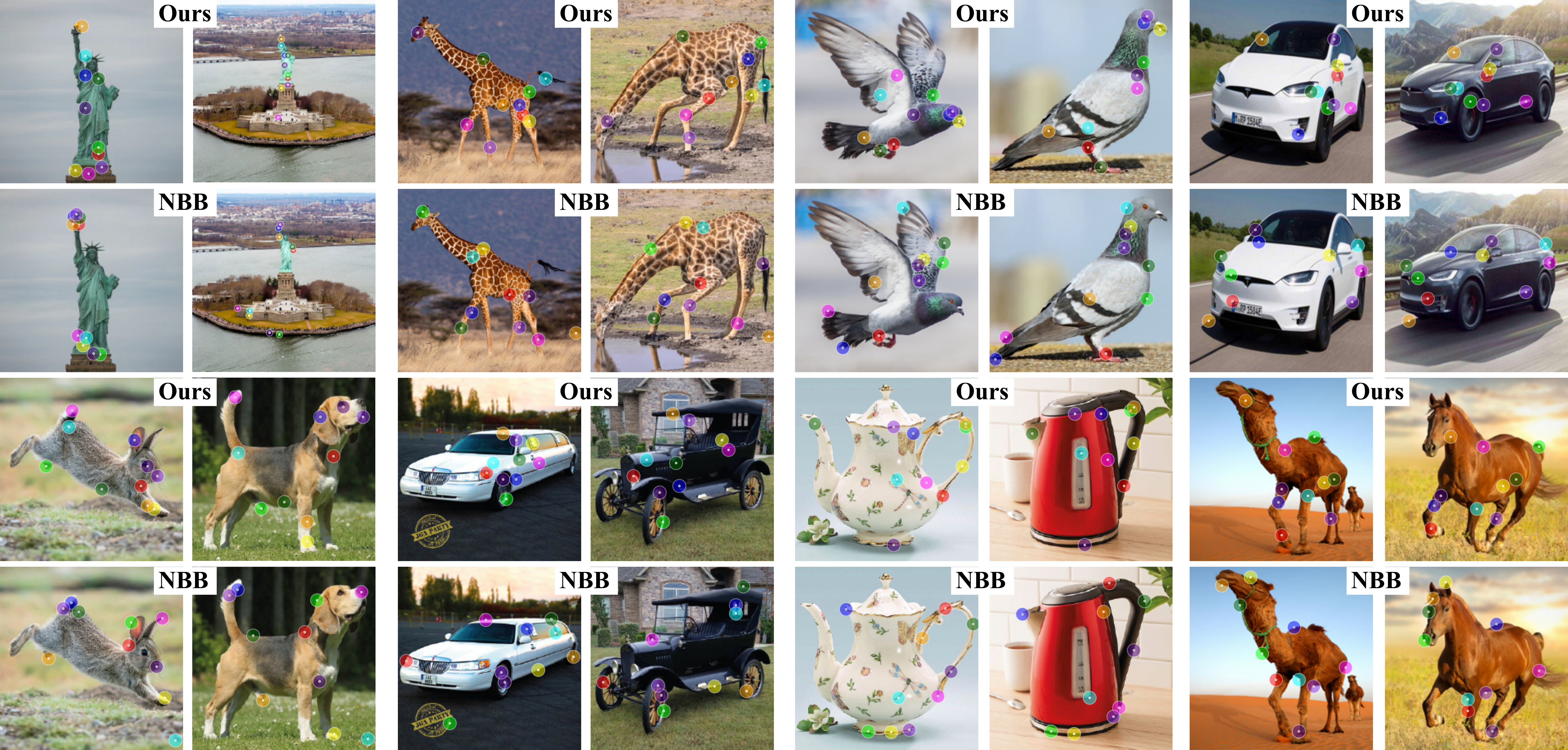}
    \caption{\textit{Correspondences Comparison to NBB~\cite{aberman2018neural}:} On intra-class (top-row) and inter-class (bottom-row) scenarios. Our method is more robust to appearance, pose and scale variations. Full size results are available in the SM.} 
    \label{fig:nbb_comp}
\end{figure}

\myparagraph{Qualitative Results.} 
We test our method on numerous pairs, compared with the VGG-based method, NBB~\cite{aberman2018neural}. 
Figure~\ref{fig:nbb_comp} shows our results are more robust to changes of appearance, pose and scale on both intra- and inter-class pairs.

\myparagraph{Quantitative Evaluation.}
We evaluate on 360 random Spair71k~\cite{min2019spair} pairs, and measure performance by Percentage of Correct Keypoint (PCK) - a predicted keypoint is considered correct if it lies within a $\alpha\cdot\max(h, w)$ radius from the annotated keypoint, where $(h, w)$ is the image size. We modify our method to match this evaluation protocol to compute the binned descriptors for the given keypoints in the source image, and find their nearest-neighbors in the target image. We compare to NBB\cite{aberman2018neural} (VGG19-based) and CATs~\cite{cat} (ResNet101-based). Table \ref{tab:correspondence_spair_table} shows that our method outperforms NBB \emph{by a large margin}, and closes the gap towards the \emph{supervised} CATs~\cite{cat}.

\begin{table}[!t]
    \centering
    \begin{tabular}{l| c| c| c| c| c| c| c| c| c| c}
    \hline
    \multirow{2}{*}{Method} & \multicolumn{4}{c|}{Layer 9} & \multicolumn{4}{c|}{Layer 11} & \multirow{2}{*}{NBB~\cite{aberman2018neural}} & \multirow{2}{*}{Supervised~\cite{cat}} \\
    \cline{2-9}
     & key & query & value & token & key & query & value & token &  & \\
    \hline
    with bins & \underline{56.48} & 54.96 & 52.33 & 56.03 & 53.45 & 52.35 & 49.37 & 50.34 & \multirow{2}{*}{26.98} & \multirow{2}{*}{\textbf{61.43}} \\
    without bins & 52.27 & 49.35 & 43.97 & 50.14 & 47.08 & 42.64 & 41.56 & 46.09 &  &  \\
    \hline
    \end{tabular}
    \caption{\textit{Correspondence Evaluation on Spair71k:} We randomly sample 20 image pairs per category, and report the mean PCK across all categories ($\alpha = 0.1$); higher is better. We include a recent supervised method~\cite{cat} for reference.} 

    \label{tab:correspondence_spair_table} 
\end{table}

\myparagraph{Ablations.}
We ablate our method using on different facets and layers, with and without binning. 
Table \ref{tab:correspondence_spair_table} empirically corroborates our observation of keys being a better than other facets, and that features from earlier layers are more sensitive to their position in the image. The correspondence task benefits from these intermediate features more than plainly using the deepest features (Sec.~\ref{sec:analysis}). 

\section{Conclusion}
 We provided new empirical observations on the internal features learned by ViTs under different supervisions, and harnessed them for several real-world vision tasks. We demonstrated the power of these observations by applying only lightweight zero-shot methodologies to these features, and still achieving competitive results to state-of-the-art supervised methods. 
 We also presented new capabilities of part co-segmentation across classes, and on domains that lack available training sets.
 We believe that our results hold great promise for considering deep ViT features as an alternative to deep CNN features.

\paragraph{Acknowledgments:} We thank Miki Rubinstein, Meirav Galun, Kfir Aberman and Niv Haim for their insightful comments and discussion.
This project received funding from the Israeli Science Foundation (grant 2303/20), and the Carolito Stiftung. Dr Bagon is a Robin Chemers Neustein Artificial Intelligence Fellow.

\bibliographystyle{splncs04}
\bibliography{egbib}
\end{document}